\documentclass[10pt,conference]{IEEEtran}
\IEEEoverridecommandlockouts
\usepackage{cite}
\usepackage{tabularx}
\usepackage{mathrsfs, amsmath,amssymb,amsfonts}
\usepackage{graphicx}
\usepackage{textcomp}
\usepackage[dvipsnames,svgnames,table]{xcolor}
\def\BibTeX{{\rm B\kern-.05em{\sc i\kern-.025em b}\kern-.08em
    T\kern-.1667em\lower.7ex\hbox{E}\kern-.125emX}}

\usepackage{lipsum}

\usepackage[utf8]{inputenc}
\usepackage[T1]{fontenc}

\usepackage[abbreviations]{foreign}
\usepackage{algorithm}
\usepackage{algpseudocode}

\usepackage{xspace}
\usepackage{siunitx}
\usepackage{listings}
\usepackage{subcaption}
\usepackage{textcomp}
\usepackage{booktabs}

\usepackage{url}
\usepackage{pifont}
\usepackage[colorlinks=true,linkcolor=blue,anchorcolor=blue,citecolor=blue,filecolor=black,menucolor=black,runcolor=black,urlcolor=blue]{hyperref}

\usepackage[inline]{enumitem}
\setlist{noitemsep,topsep=0pt,parsep=0pt,partopsep=0pt}

\usepackage[capitalise]{cleveref}

\usepackage{tikz}
\usetikzlibrary{positioning, arrows.meta}
\usetikzlibrary{matrix}

\definecolor{lightcolor}{rgb}{0,0.5,1}

\usepackage{fontawesome5}
\newboolean{showcomments}
\setboolean{showcomments}{false}
\ifthenelse{\boolean{showcomments}}
{ \newcommand{\mynote}[3]{
		\fbox{\bfseries\sffamily\scriptsize#1}
		{\small$\blacktriangleright$\textsf{\emph{\color{#3}{#2}}}$\blacktriangleleft$}}}
{ \newcommand{\mynote}[3]{}}

\definecolor{darkgreen}{rgb}{0.3,0.5,0.3}
\definecolor{darkblue}{rgb}{0.3,0.3,0.5}
\definecolor{darkred}{rgb}{0.5,0.3,0.3}

\newcounter{numobserv} 
\definecolor{beaublue}{rgb}{0.88, 0.93, 0.93}
\usepackage[most]{tcolorbox}
\colorlet{shadecolor}{beaublue}
\tcbset{
	colback=beaublue,
	boxrule=0.5pt,
	boxsep=0pt,
	left=1pt,
	right=1pt,
	top=1pt,
	bottom=1pt,
	sharp corners,
	colframe=white,
}

\newcommand{\cmark}{\textcolor{NavyBlue}{\ding{51}}}%
\newcommand{\xmark}{\color{BrickRed}{\ding{55}}}%

\usepackage{flushend}

\makeatletter
\let\origsection\section
\renewcommand\section{\@ifstar{\starsection}{\nostarsection}}
\newcommand\nostarsection[1]{\sectionprelude\origsection{#1}\sectionpostlude}
\newcommand\starsection[1]{\sectionprelude\origsection*{#1}\sectionpostlude}
\newcommand\sectionprelude{\vspace{2pt}}
\newcommand\sectionpostlude{\vspace{2pt}}
\let\origsubsection\subsection
\renewcommand\subsection{\@ifstar{\starsubsection}{\nostarsubsection}}
\newcommand\nostarsubsection[1]{\subsectionprelude\origsubsection{#1}\subsectionpostlude}
\newcommand\starsubsection[1]{\subsectionprelude\origsubsection*{#1}\subsectionpostlude}
\newcommand\subsectionprelude{\vspace{2pt}}
\newcommand\subsectionpostlude{\vspace{2pt}}
\g@addto@macro\normalsize{%
  \setlength\abovedisplayskip{2pt}
  \setlength\belowdisplayskip{2pt}
  \setlength\abovedisplayshortskip{2pt}
  \setlength\belowdisplayshortskip{2pt}
  \setlength{\floatsep}{3pt}
  \setlength{\textfloatsep}{3pt}
  \setlength{\intextsep}{2pt}
  \setlength{\dblfloatsep}{3pt}
  \setlength{\dbltextfloatsep}{3pt}
}
\makeatother

\usepackage[export]{adjustbox}
\usepackage{eso-pic}

\begin{document}


\title{On the Cost of Model-Serving Frameworks:\\An Experimental Evaluation}

\author{\IEEEauthorblockN{Pasquale De Rosa}
\IEEEauthorblockA{\textit{University of Neuch\^{a}tel}\\
Neuch\^{a}tel, Switzerland \\
pasquale.derosa@unine.ch} \and 
\IEEEauthorblockN{Yérom-David Bromberg}
\IEEEauthorblockA{\textit{University of Rennes, CNRS, INRIA, IRISA}\\
Rennes, France \\
david.bromberg@irisa.fr} \and
\IEEEauthorblockN{Pascal Felber}
\IEEEauthorblockA{\textit{University of Neuch\^{a}tel}\\
Neuch\^{a}tel, Switzerland \\
pascal.felber@unine.ch} \and
\IEEEauthorblockN{Djob Mvondo}
\IEEEauthorblockA{\textit{University of Rennes, CNRS, INRIA, IRISA}\\
Rennes, France \\
barbe-thystere.mvondo-djob@inria.fr} \and
\IEEEauthorblockN{Valerio Schiavoni}
\IEEEauthorblockA{\textit{University of Neuch\^{a}tel}\\
Neuch\^{a}tel, Switzerland \\
valerio.schiavoni@unine.ch}}

\maketitle

\thispagestyle{plain}
\pagestyle{plain}
%

\def\confname{12th IEEE International Conference on Cloud Engineering (IC2E'24)}
\def\confyear{2024}
\def\confdoi{XXX}

\definecolor{yellowPaper}{HTML}{fff8ae}
\AddToShipoutPictureFG*{%
  \AtTextUpperLeft{%
    \adjustbox{raise=3pt}{
    \begin{tcolorbox}[width=1\textwidth,colback=yellowPaper,enhanced,frame hidden,sharp corners]  
        \centering\scriptsize
        \copyright~\confyear\ 	
         by the Institute of Electrical and Electronics Engineers (IEEE). Personal use of this material is permitted. Permission from IEEE must be obtained for all other uses, in any current or future media, including reprinting/republishing this material for advertising or promotional purposes, creating new collective works, for resale or redistribution to servers or lists, or
         reuse of any copyrighted component of this work in other works.
        This is the author's version of the work.
        The final authenticated version is available online at \href{https://doi.org/10.1109/IC2E61754.2024.00032}{https://doi.org/10.1109/IC2E61754.2024.00032} 
        and has been published in the proceedings of the 
        \confname.
     \end{tcolorbox}} 
  }%
}%

\hypersetup{
    pdftitle={\copyright~\confyear\  Copyright 2023 by the Association for Computing Machinery, Inc. (ACM). Permission to make digital or hard copies of portions of this work for personal or classroom use is granted without fee provided that the copies are not made or distributed for profit or commercial advantage and that copies bear this notice and the full citation on the first page in print or the first screen in digital media. Copyrights for components of this work owned by others than ACM must be honored. Abstracting with credit is permitted.
    	This is the author's version of the work.
    	The final authenticated version is available online at \href{https://doi.org/10.1145/3583678.3596888}{https://doi.org/10.1145/3583678.3596888} 
    	and has been published in the proceedings of the 
    	\confname.}
}

\begin{abstract}
In machine learning (ML), the inference phase is the process of applying pre-trained models to new, unseen data with the objective of making predictions. 
During the inference phase, end-users interact with ML services to gain insights, recommendations, or actions based on the input data.  
For this reason, serving strategies are nowadays crucial for deploying and managing models in production environments effectively. These strategies ensure that models are available, scalable, reliable, and performant for real-world applications, such as time series forecasting, image classification, natural language processing, and so on. In this paper, we evaluate the performances of five widely-used model serving frameworks (TensorFlow Serving, TorchServe, MLServer, MLflow, and BentoML) under four different scenarios (malware detection, cryptocoin prices forecasting, image classification, and sentiment analysis). We demonstrate that TensorFlow Serving is able to outperform all the other frameworks in serving deep learning (DL) models. Moreover, we show that DL-specific frameworks (TensorFlow Serving and TorchServe) display significantly lower latencies than the three general-purpose ML frameworks (BentoML, MLFlow, and MLServer).
\end{abstract}


\section{Introduction}
\label{sec:intro}
Machine learning systems (ML) focus on algorithms and models that enable computers to learn from data and make predictions or decisions. 
Common applications of ML include price predictions of financial assets~\cite{appl1, appl2}, recommendation systems~\cite{appl3, appl4}, object detection~\cite{appl5, appl6}, speech recognition~\cite{appl7, appl8}, and more. 
A typical ML processing pipeline consists broadly of two phases.
The training phase, \ie the process of building a model from data, is generally computationally expensive.
It is executed mostly by complex model structures on large datasets~\cite{clipper}. 
State-of-the-art neural network architectures (such as recurrent neural networks - RNNs \cite{gru, lstm}, and convolutional neural networks - CNNs \cite{cnn}), despite their excellent recognition accuracy, come at very high computational costs induced by their training phase~\cite{rnnvisual}. 
More specifically, CNNs, typically used in image classification tasks\cite{cnn2}, can require several days to train on multiple GPUs even with downsampled input images~\cite{convnetcost}. 

The inference (or prediction) phase is the process of building predictions from the trained model.
It is generally considered less expensive since it does not require multiple iterations over batched samples of the dataset~\cite{clipper}. 
This phase consists of a single forward pass\footnote{The process of propagating input data through the ML model to produce an output.} over the test set.
If considering models fitted with a very high number of parameters, \eg neural networks, it is generally smaller in size than the data used for training~\cite{joseph}. 

While the training phase is time-consuming, 
the inference should normally execute in \emph{almost} real-time, as is typically embedded in user-facing applications that provide low-latency predictions at scale~\cite{clipper}.
To design these user-facing prediction services, the Machine Learning as a Service (\emph{MLaaS})~\cite{ribeiro2015mlaas} paradigm allows to automatically train, test, and deploy ML models, handling the entire model lifecycle.

Nowadays, it is common for major cloud providers to enhance their products with functionalities enabling users to deploy ML projects~\cite{cloud1, mlaas}. 
Cloud providers are generally a good solution to develop \emph{MLaaS} platforms because they control a vast volume of data and provide a massive computational power, both instrumental to develop complex ML services for the large audience~\cite{cloud1, mlaas}. 


However, while relying on \emph{MLaaS} platforms is considered the most practical solution, not involving structural costs traditionally linked to in-premises infrastructures (\ie ownership and maintenance of machines, in-house IT teams), it also comes at a steep price. 
For example, a standard instance with 2 vCPUs and 8 GiB RAM on AWS Sagemaker~\cite{aws} costs about \$0.10/h, on Azure Machine Learning~\cite{azure} costs \$0.096/h and on Google AI Platform~\cite{googleai} up to \$0.10/h.\footnote{Prices for the East US region as of May 2024.} 
Considering how production-ready deployments require much higher computing resources, such costs will increase considerably.
Bigger cloud instances with 64 vCPUs and 256 GiB RAM, capable of efficiently executing optimized ML tasks, cost \$968'680/year (AWS), \$807'321/year (Azure), and \$856'728/year (Google). 

In addition, outsourcing a sensitive model to untrustworthy cloud providers comes with serious drawbacks, both from a technical and jurisdictional perspective \cite{conf1, conf2}. 

To sidestep the mentioned drawbacks, one can consider an on-premise deployment of an \emph{MLaaS} service over its private cloud infrastructure, typically built on top of an open-source and easy-to-access ML serving framework. 
Several frameworks allow providers to design scalable and efficient ML inference services and expose them via public-facing APIs.




However, the systematic evaluation of different model-serving strategies remains an open challenge despite end-users increasing demand for real-time ML predictions. 

This paper aims to fill this existing gap by benchmarking five state-of-the-art open-source model serving frameworks, namely \sloppy{TensorFlow-Serving}~\cite{tfs,olston2017tensorflow}, TorchServe~\cite{ts}, MLServer~\cite{mlserver}, MLflow~\cite{mlflow,zaharia2018accelerating} and BentoML~\cite{bentoml}. 

The reason why we select these frameworks is to have a representative of both DL-specific solutions (\ie TensorFlow Serving and TorchServe), the \emph{de-facto} baseline solutions in serving models built, respectively, with TensorFlow \cite{tensorflow2015-whitepaper} and PyTorch \cite{torch}, and general-purpose frameworks for ML deployment (\ie MLServer, MLflow and BentoML).

In this study we focus on the design and evaluation of different serving runtimes, packaging model code and artifacts into containers to build APIs optimized for model inference. We leave the discussion of end-to-end serving platforms based on Kubernetes \cite{kubernetes}, like KServe \cite{kserve} and Kubeflow \cite{kubeflow}, to future extensions of this work.

To thoroughly evaluate such frameworks, we consider four different and diverse application scenarios: \emph{(i)} malware detection, \emph{(ii)} cryptocoin prices forecasting, \emph{(iii)} image classification, and \emph{(vi)} sentiment analysis. 
These scenarios represent four realistic use cases from real-world business workloads \cite{realworld1, realworld2, realworld3, realworld4}, where  ML inference services are implemented. However, the flexible nature of our study makes its results generalizable to multiple business applications, in addition to these ones.

Moreover, these scenarios allow our study to offer a wide range of real-world ML cases: regressions for the price forecasting use-case, binary classification (for malware detection and sentiment analysis), and multiclass classification (on images), ensuring the robustness and generalizability of our findings.

\begin{tcolorbox}	
	\emph{Our results show that TensorFlow Serving consistently outperforms all the other frameworks in serving deep learning (DL) models.
The two DL-specific frameworks (TensorFlow Serving and TorchServe) display lower latencies than the remaining general-purpose frameworks.}	
\end{tcolorbox}

The main contributions of this paper are:

\begin{itemize}
	\item The systematic analysis of five state-of-the-art open source model serving frameworks on four real-world application scenarios;
	\item The identification of the best platform among the considered ones in terms of model serving latencies;
	\item The assessment of the best performing approach to DL model serving between DL-specific frameworks (TensorFlow Serving and TorchServe) and general-purpose ML frameworks (MLflow, MLServer, BentoML). 
\end{itemize}



We follow an open science approach, releasing code and pre-trained models to the research community at~\url{https://zenodo.org/records/11499446}.

\textbf{Roadmap.} \S\ref{sec:background} describes the details of the model serving frameworks that we consider in this paper. 
\S\ref{sec:scenarios} provides an overview of the considered application scenarios. 
\S\ref{sec:dataset} characterizes our datasets.
\S\ref{sec:eval} discusses the experimental evaluation, while in \S\ref{sec:discussion} we present our results.
We cover related work in
\S\ref{sec:rw}, before concluding with our future work in \S\ref{sec:conclusion}.

\begin{table*}[!t]
	\hspace*{-9.5cm}
	\scriptsize
	\centering
	\setlength{\tabcolsep}{2pt}
		\rowcolors{1}{gray!10}{gray!0}
		\begin{tabularx}{\columnwidth}{l|rrrr|r|r|r}
			\rowcolor{gray!50}
			\textbf{Framework} & \textbf{ML} & \textbf{DL} & \textbf{CPU} & \textbf{GPU} & \textbf{Supported (ML)} & \textbf{Supported (DL)} & \textbf{Model Formats} \\
			\rowcolor{gray!10}
			TF-Serving   &  \xmark &  \cmark & \cmark & \cmark & - & TensorFlow, HuggingFace, Keras & SavedModel, TFLite, TF.js \\
			\rowcolor{gray!0}
			TorchServe   & \xmark &  \cmark &  \cmark &  \cmark & - & PyTorch, HuggingFace & MAR, PyTorch, ONNX \\
			\rowcolor{gray!10}
			MLServer   &  \cmark  &  \cmark &  \cmark &  \cmark & Scikit-Learn, XGBoost, LightGBM, CatBoost & TensorFlow, PyTorch, HuggingFace, Keras & SaveModel, PyTorch, Keras, ONNX, Pickle, Joblib \\
			\rowcolor{gray!0}
			MLflow   & \cmark & \cmark &  \cmark &  \cmark & Scikit-Learn, XGBoost, LightGBM, CatBoost & TensorFlow, PyTorch, HuggingFace, Keras & SaveModel, PyTorch, Keras, ONNX, Pickle, Joblib \\
			\rowcolor{gray!10}
			BentoML   & \cmark & \cmark & \cmark & \cmark & Scikit-Learn, XGBoost, LightGBM, CatBoost & TensorFlow, PyTorch, HuggingFace, Keras & SaveModel, PyTorch, Keras, ONNX, Pickle, Joblib \\
		\end{tabularx}
	\caption{\label{tab:serving-frameworks}Overview of the model serving frameworks (supported ML/DL libraries and model formats are shortlisted from the official docs).}
	

\end{table*}

\section{Model Serving Frameworks}
\label{sec:background}
In this section, we provide insights on the selected model serving frameworks, \,  e.g., TensorFlow Serving, TorchServe, MLServer, MLflow, and BentoML.

\autoref{tab:serving-frameworks} provides a comprehensive overview of these serving frameworks, highlighting their CPU/GPU support capability and the pool of ML/DL libraries they handle. 
We further detail such frameworks in the rest of this section.

\textbf{TensorFlow Serving}~\cite{tfs} is a flexible, high-performance model-serving system for ML and deep learning (DL) models, typically designed for production environments. 
TensorFlow Serving provides a native integration with TensorFlow DL models.

Pretrained models are serialized in a specific format (\ie \texttt{SavedModel}), that contains a complete TensorFlow program, comprising of the trained parameters, the variables and the computation. 
Models can be saved and loaded in the \texttt{SavedModel} format using either a low-level \texttt{tf.saved\_model} API or an higher-level \texttt{tf.keras.Model} API.

Once the model has been stored in the \texttt{SavedModel} and correctly loaded in the registry\footnote{A centralized repository that stores versioned models, metadata, and lineage information.}, TensorFlow Serving will accept inference requests, typically over REST or gRPC interfaces.

\textbf{TorchServe}~\cite{ts} is a tool that streamlines the deployment of PyTorch models in production environments, offering scalability, flexibility, and performance optimization.
A PyTorch model (\texttt{.pth}/\texttt{.pt}) differs from a TensorFlow \texttt{SavedModel}; while the first offers modular saving options (TorchScript and state dictionary), the latter provides an all-in-one encapsulation.
Moreover, PyTorch models can be deployed using TorchScript \cite{torchscript}, whereas \texttt{SavedModel} supports a wider range of deployment environments (\eg, TensorFlow Serving, TensorFlow.js \cite{tfjs}, TensorFlow Lite \cite{tflite}).
Being primarily focused on serving PyTorch models, it does not provide out-of-the-box support for other ML frameworks.
Its popularity among practitioners and ML enthusiasts is high, and popular model-hosting sites, such as HuggingFace\footnote{\url{https://huggingface.co/}} and their  Hugging FaceTransformers, which are typically developed in PyTorch, can also be served using TorchServe~\cite{hfts}. 

To serve a model with TorchServe, it must be archived as a MAR file using the Torch Model Archiver \cite{mar}. 
TorchServe can automatically scale the total number of backend workers serving the model to the number equal to the available vCPUs or GPUs on the underlying computing node.

TorchServe also allows it to serve multiple models simultaneously within a single server instance. 
This can be instrumental in cases where a user requests more than one prediction at a time from different models.

Finally, TorchServe allows users to define custom handlers for preprocessing and postprocessing model inputs and outputs, allowing them to define flexible services that can adapt to specific requirements and applications.

\textbf{MLServer}~\cite{mlserver} is a scalable and flexible framework optimized for serving ML models in production environments. 
MLServer provides out-of-the-box support for several ML frameworks (\eg Scikit-learn \cite{sklearn}, XGBoost \cite{xgboost}, LightGBM \cite{lightgbm}, CatBoost \cite{catboost}), as well as for DL frameworks like TensorFlow and PyTorch. 
This flexibility enables users to deploy a large set of ML models regardless of the framework they were originally trained in.

MLServer is highly scalable and capable of handling many inference requests concurrently. 
Moreover, it allows users to serve models in custom environments specifically designed to match the runtime requirements of different models.

MLServer implements a RESTful API endpoint for inference to serve predictions at scale using the models stored in the registry.

\textbf{MLflow}~\cite{mlflow} is a comprehensive platform for managing the entire ML pipeline, from model training to experimentation, tracking/logging and deployment. 
Being general-purpose, it allows to serve ML and DL models trained in various frameworks, like TensorFlow, PyTorch, Scikit-learn \etc.

MLflow provides tracking and logging features for the deployed models, enabling real-time model performance analysis. 
Moreover, it allows you to define custom pre- and post-processing logic to be integrated into the model inference service. 

MLflow relies on Flask~\cite{flask} to serve the inference endpoint, but integration with MLServer or other alternatives is possible. 

\textbf{BentoML} \cite{bentoml} is a unified model-serving library for building scalable AI applications with Python. It includes tools for serving optimization, model packaging, and production deployment.
Like MLServer and MLflow, BentoML is a general-purpose framework that supports a wide range of ML libraries, including TensorFlow, PyTorch, Scikit-learn, XGBoost, LightGBM, and CatBoost.

BentoML allows the packaging of pre-trained models and all the necessary requirements into a comprehensive, ready-to-deploy format named Bento. A Bento includes all the libraries, configurations, and pre-and post-processing logic needed for the model to produce inferences at scale correctly.

\subsection{Models: formats and conversions}
As described before, TorchServe is not natively intended to handle deploying models not designed in PyTorch. Therefore, since our models were developed using TensorFlow (in \texttt{SavedModel}), we needed to convert them into PyTorch format (\texttt{.pth}/\texttt{.pt}).

However, this task was challenging since the two formats proved to be very different both in structure and purpose (for a more comprehensive overview, refer to \autoref{sec:background}).

One common conversion method combines TensorFlow's support for exporting models to the ONNX format \cite{onnx} and PyTorch's ability to import ONNX models. 
The ONNX (Open Neural Network Exchange) model format is an open-source interchange\footnote{A file format designed to facilitate the exchange of models between different platforms, providing a standardized way to represent such models.} format for representing DL models. It aims to provide a standard way to allow models to be trained in one framework and deployed in another. It enables interoperability between different DL frameworks, just like TensorFlow and PyTorch: DL models trained in one framework can be exported to ONNX and imported into another framework for inference (or further training).

Undoubtedly, the cross-framework conversion of a model might generate a potential threat to validity. However, given the current state of the art, this approach is the only feasible way to allow a comparison between the two most commonly used DL frameworks, TensorFlow and PyTorch, and hereby enrich the quality of our study.

In our work, we leveraged two Python libraries written on top of ONNX: TF2ONNX \cite{tf2onnx}, which allows us to convert TensorFlow models to ONNX, and ONNX2Torch \cite{onnx2torch}, an ONNX to PyTorch converter.


\section{Motivating Scenarios}\label{sec:scenarios}
We describe the four scenarios we consider best to conduct our experimental evaluation of the various model-serving frameworks.
These scenarios capture different application contexts: \emph{(i)} malware detection, \emph{(ii)} cryptocoin price forecasting, \emph{(iii)} image classification, and \emph{(iv)} sentiment analysis.

\textbf{Malware detection} on Android APKs (Android application packages) is the analysis of the file features to determine whether it contains malicious code or behavior. 
Nowadays, it is common to adopt ML and DL algorithms to perform this classification task~\cite{md1, md2, md3, drebin}.

Android malware detection approaches can be mostly divided into static and dynamic analyses. 
The static analysis extracts the features from an APK without running them into a device or Android emulator.
Instead, the dynamic analysis extracts the features by running them on an Android emulator or device~\cite{md1, md2}.

In our evaluation, we rely on a static analysis approach. Our choice is motivated by the fact that, on malware detection tasks using ML techniques, dynamic analysis proved to be less effective, thus leading to a minor recognition accuracy if compared with a static approach \cite{staticvsdynamic}.

The first step in the malware detection process is to extract relevant features from the APK file. 
These features can include permissions requested by the application, API calls, code structure, metadata, resource files, \etc. 
To extract features from the original APKs, we leveraged Drebin~\cite{drebin}, a lightweight static analyzer for Android applications. 
Then, we generated a vectorial representation of each application with values in $\{0,1\}$, where 1 means that a specific feature is present in the APK and 0 otherwise.

Therefore, we used a simple Deep Neural Network (DNN) with two hidden layers of 200 units and a final sigmoid layer to recognize a malicious/benign APK from the previously generated vectorial representation. 
We trained our DNN model on a subsample of the CICMalDroid 2020 dataset~\cite{mald20}, containing the 10\% of the observations and following the same distribution between malicious and benign examples of the original data.

\textbf{Cryptocoin price forecasting} is the analysis of co-movements and correlation patterns in the cryptocurrency market to forecast the price trends for a given coin. 
History has shown the extreme volatility of such trading prices. Reasons for such instability include the lack of adequate regulation, the inherent speculative nature of cryptocurrencies, and the lack of a governmental or institutional guarantor. 
However, recent studies~\cite{crypto1, crypto2, crypto3} demonstrated the possibility of forecasting these particular time series using ML and DL models.

In our scenario, we leveraged the price series trends of a set of 8 highly-correlated coins (\ie ADA, BNB, DOGE, ETH, LTC, SOL, XLM, XRP) to forecast the hourly price trends of Bitcoin (BTC) over a user-specified number of time steps.

We used a DNN with two hidden layers of 200 units and a final ReLu layer to forecast the BTC price using the correlated feature coins' series as feature variables to achieve this target. 
To estimate the feature coins' series over the desired time frame, we adopted a Moving Average (MA) over a sliding window of the last 24 hours' observations from the validation set. 
We trained our DNN model on a dataset gathered from Binance~\cite{binance} (a popular cryptocoin exchange) containing the closing prices of the analyzed cryptocoins between within 19 days, \ie 01-01-2024 00:00:00 and 19-04-2024 23:00:00.

\textbf{Image classification} is a widely analyzed task in ML.
It consists of categorizing images into predefined classes or labels. 
Crucial applications of image recognition include, for example, medical diagnosis~\cite{image1} and object detection~\cite{image2}.

In our scenario, as a classification model, we used  MobileNetV2~\cite{mobilenetv2}, \ie a Convolutional Neural Network (CNN) based on an inverted residual structure with connections between the bottleneck layers.
The model has been trained on the ImageNet 2012 dataset~\cite{imagenet}, first presented in the ImageNet Large Scale Visual Recognition Challenge 2012 (ILSVRC 2012) \cite{ilsvrc12}. 
The dataset spans \num{1000} object classes and contains \num{1281167} training images, \num{50000} validation images, and \num{100000} test images. 

The pre-trained model, containing the loaded ImageNet weights, was retrieved from Keras Applications \cite{kerasa}. 
It can recognize image input data of shape (224, 224, 3).

\textbf{Sentiment Analysis}, or opinion mining, is a natural language processing (NLP) application that involves the analysis of opinions, attitudes, emotions, and sentiments expressed in textual data (and, more recently, in other types of media).
Common real-world applications of sentiment analysis involve the study of opinions on social media~\cite{sentiment1} and the analysis of customer feedback from reviews~\cite{sentiment2}.

In this scenario, we used a Convolutional Neural Network (CNN) with an initial Embedding layer, two convolutional layers, and a final sigmoid layer to recognize a negative/positive sentiment from a tokenized\footnote{Tokenizers convert raw string inputs into integer inputs suitable for the Embedding layer.} input sentence. 

To tokenize the input data, we cleaned it by removing punctuations and stopwords and applying a lemmatization\footnote{The process of reducing words to their base, known as the lemma.}.

We trained our model on the IMDb 50K dataset~\cite{imdb}, containing \num{50000} movie reviews for natural language processing and text analytics applications.

\autoref{tab:scenarios} provides a comprehensive overview of these scenarios, highlighting the combination of dataset and models used in each.

\begin{table}[!t]
    	\caption{\label{tab:scenarios}Overview of the motivating scenarios with datasets and models used.}
	\centering
    \footnotesize
	\setlength{\tabcolsep}{10pt}
	\rowcolors{1}{gray!10}{gray!0}
	\begin{tabularx}{\columnwidth}{lc}
		\rowcolor{gray!50}
        \textbf{UC1: Malware Analysis} &   \\		
		\rowcolor{gray!10}
		\textbf{Dataset}  &  CICMalDroid 2020 \cite{mald20} (10\%) \\
		\rowcolor{gray!0}
		\textbf{Model}   &  DNN + Drebin \cite{drebin}   \\
        
		\rowcolor{gray!50}
		\textbf{UC2: Cryptocoin} &  \\
		\rowcolor{gray!10}
		\textbf{Dataset} & Binance \cite{binance}   \\
		\rowcolor{gray!0}
		\textbf{Model}   & DNN + MA \\

		\rowcolor{gray!50}
		\textbf{UC3: Image classification} &  \\
		\rowcolor{gray!10}
		\textbf{Dataset} & ImageNet 2012 \cite{imagenet}  \\
		\rowcolor{gray!0}
		\textbf{Model}   & MobileNetV2 \cite{mobilenetv2}\\

		\rowcolor{gray!50}
		\textbf{UC4: Sentiment analysis} &  \\
		\rowcolor{gray!10}
		\textbf{Dataset}   & IMDb 50K \cite{imdb}  \\
		\rowcolor{gray!0}
		\textbf{Model}  &  Embedding + CNN \cite{cnn}\\
            
	\end{tabularx}

	
\end{table}

\section{Datasets}\label{sec:dataset}
To execute our experimental evaluation, we rely on four real-world datasets. 
Each is used in the considered scenarios (\S\ref{sec:scenarios}).
The datasets are: CICMalDroid 2020~\cite{mald20}, Binance~\cite{binance}, ImageNet~\cite{imagenet} and IMDb 50K~\cite{imdb}.

\textbf{CICMalDroid 2020.} 
This dataset is an Android malware dataset with more than \num{17341} samples, collected from several sources, including VirusTotal~\cite{virustotal} or Contagio~\cite{contagio}, from December 2017 to December 2018.

The APKs dataset spans five categories: adware, Banking malware, SMS malware, Riskware, and Benign. 
More specifically, Adware refers to advertising material that hides inside apps infected by malware. Banking malware is designed to access the user’s online banking accounts. SMS malware exploits the SMS service to conduct attacks. Riskware is a legitimate program that can cause damage if used maliciously, and Benign is a nonmalicious, safe application.

We gathered $\approx 10\%$ of the total samples (1727), following the original distribution of the original data between Adware (8.8\%), Banking malware (14.5\%), SMS malware (27.9\%), Riskware (25.2\%) and Benign samples (23.6\%). 

In terms of sample size, the largest benign APK in our dataset is 47MB, while the smallest is 48KB; for malicious samples, instead, the biggest APK size is 41MB while the smallest is 8KB.

\textbf{Binance.} 
In this dataset, we gathered time series from Binance, a popular crypto coin exchange.
The time series provides high resolution data points (\ie, one sampling per hour). 
It includes the closing price\footnote{The last price at which the cryptocoin traded during the hour.} records for 9 top-traded cryptocoins (\eg ADA, BNB, BTC, DOGE, ETH, LTC, SOL, XLM, XRP). 
The entire dataset consists of \num{23544} observations from 01-01-2024 00:00:00 to 19-04-2024 23:00:00. 
Each coin's time series includes \num{2616} steps.

\textbf{ImageNet}.
We use ImageNet~\cite{imagenet}, a well-known dataset and image database organized according to the WordNet~\cite{wordnet} hierarchy. WordNet is a large lexical database for the English language, combining a dictionary and thesaurus. In WordNet, nouns, verbs, adjectives, and adverbs are grouped into cognitive synonym rings (synsets). In the ImageNet database, each image is annotated using WordNet synsets, making it one of the largest resources available for training DL models in computer vision tasks. 
While the original dataset contains over 14 million images, its most popular version, \eg ILSVRC (ImageNet Large Scale Visual Recognition Challenge) 2012~\cite{ilsvrc12}, spans \num{1000} object classes and contains \num{1281167} training images, \num{50000} validation images and \num{100000} test images.

\textbf{IMDb 50K} is a popular dataset collected from IMDb (Internet Movie Database)~\cite{imdb}. 
This dataset, commonly used for binary sentiment classification (positive/negative), contains \num{25 000} highly polar movie reviews for training and \num{25 000} for testing purposes.

\section{Evaluation}
\label{sec:eval}
In this section, we provide a detailed overview of the experimental evaluation of the selected model-serving frameworks (TensorFlow Serving, TorchServe, MLServer, MLflow, and BentoML) in each of the four considered scenarios. 
We aim to evaluate different strategies of the DL model serving under realistic circumstances and assess which framework is most efficient in terms of observed latencies.
Specifically, our experimental evaluation intends to answer the following questions:
\begin{itemize}
	\item \emph{Q1:} Are there significative differences in the observed latencies for the considered model serving platforms?
    \item \emph{Q2:}Can significative differences be observed in the latencies for different sizes of the input payload?
    \item \emph{Q3:} Which model serving framework displays the most efficient performance on DL pipelines in terms of low latencies?
    \item \emph{Q4:} Are DL-specific frameworks (TensorFlow Serving and TorchServe) capable of outperforming the general-purpose frameworks (MLflow, MLServer, and BentoML) on DL deployment tasks?
\end{itemize}
We support experimental reproducibility and release our code, datasets, and instructions to reproduce our experiments at~\url{https://zenodo.org/records/11499446}.

\smallskip\noindent\textbf{Experimental setup.}
We use Ubuntu 22.04.4 LTS, 40 CPUs Intel(R) Xeon(R) Gold 5215 clocked at 2.50GHz, Linux kernel 5.15.0-105-generic, and 128 GB RAM.

We leverage Python v3.10.12, TensorFlow with TensorFlow ModelServer v2.15.0, PyTorch v2.2.2 with TorchServe 0.10.0,  BentoML v1.2.12, MLServer v1.5.0 and MLflow v2.12.1. 

We trained our models in TensorFlow. To deploy them on TorchServe, which natively supports only PyTorch pretrained formats (\texttt{.pth}), we converted the original models to PyTorch via ONNX~\cite{onnx}. 
We leveraged ONNX v1.16.0, with ONNX2Torch~\cite{onnx2torch} v1.5.14 and TF2ONNX~\cite{tf2onnx} v1.16.1.


\smallskip\noindent\textbf{Data pre-processing.}
We aim to simulate a realistic interaction between a user and an inference platform for each of the considered scenarios. 
Therefore, we use real-world testing data, to which  we apply a range of preprocessing techniques to transform the raw input into a format that can be "understood" and managed by the inference model.
A comprehensive description of the input preprocessing techniques applied in each scenario is summarized in \autoref{tab:scenarios-eval}.

\begin{enumerate}[leftmargin=*]
	\item \textbf{Scenario 1:} since the context of this scenario is to perform a malware detection on Android applications, the raw input data is an APK file (\eg, an archive compressed using the classical ). 
Using Drebin, we extract features from a given APK and confront them with the \num{30050} unique features gathered from the training set. 
Then, we generate a new vectorial representation for the APK of size \num{30050}, where each data point is either a 0 or 1, depending on whether the specific feature is present in the application.  
Therefore, the preprocessed output is a tensor\footnote{A mathematical object that generalizes scalars, vectors, and matrices to higher dimensions.} of shape  $(1, 30050)$.
	
	\item \textbf{Scenario 2:} in this scenario, the aim is to forecast a number $n$ of future hourly prices for Bitcoin, given the time series of highly-correlated \emph{altcoins}.\footnote{Altcoins are all cryptocurrencies other than Bitcoin.}
    As stated in~\S\ref{sec:background}, we forecast these series by applying a Moving Average (MA) over the last 24 prices in the validation set for each altcoin. The user-provided input is the number $n$ of future steps. 
    The resulting prediction from the DNN is generated on a preprocessed input tensor of shape $(n, 8)$, where 8 are the MA prices for the feature altcoins used in the forecasting procedure (one among ADA, BNB, DOGE, ETH, LTC, SOL, XLM or XRP).
	
	\item \textbf{Scenario 3:} in the image classification scenario, clients provide an input image in JPEG format that can be of any resolution and shape. To make it ready for the MobileNetV2 inference, we reshape it in an image $224$x$224$x$3$, that corresponds to a tensor of shape $(1, 224, 224, 3)$.\footnote{The first dimension (\ie 1) is the batch size, that for a single input is unitary.}
	
	\item \textbf{Scenario 4:} finally, in this scenario the user provides an input sentence/opinion that the DL model classifies as either positive or negative. 
To make the data inference-ready, we apply tokenization and lemmatization and remove stopwords from the original sentence. 
Then the text is converted to a sequence of integers, where each integer represents a specific word in the vocabulary. Therefore, the resulting preprocessed input is a tensor of shape $(n, 1)$, namely a list containing a sequence of the $n$ tokens extracted from the input data.
\end{enumerate}

\begin{table*}[!t]
	\hspace*{-3.5cm}
	\centering
	\setlength{\tabcolsep}{4pt}
	\setlength\extrarowheight{3pt}
	\rowcolors{1}{gray!10}{gray!0}
	\begin{tabularx}{\columnwidth}{lrrrr}
		\rowcolor{gray!50}
		& \textbf{Scenario 1} & \textbf{Scenario 2} & \textbf{Scenario 3} & \textbf{Scenario 4} \\
		\rowcolor{gray!10}
		\textbf{Original Input}   &  APK file &  Number of steps & JPEG image & Text  \\
		\rowcolor{gray!0}
		\textbf{Preprocessed Input}   &  Tensor (1, 30050) &  Tensor ($n$, 8) & Tensor (1, 224, 224, 3) & Tensor ($n$, 1)  \\
	\end{tabularx}
	\caption{\label{tab:scenarios-eval}Overview of original input and preprocessed one for each scenario.}
\end{table*}

To assess the performance of the five serving frameworks in each scenario, we rely on oha~\cite{oha}, an HTTP load generator built on Rust. 
On the client side, we set each run's duration to 1 minute, with a constant throughput of 200 requests/s. 
On the server side, we set the number of parallel threads to 40, equal to the number of CPUs available in the server hosting the inference service. 

We evaluated the latency distribution of each response for three different payload sizes: small, medium and big. Therefore, we ran a total of twelve experiments, three for each scenario. 
In scenario 1, the size of the payload corresponds to the dimension, in bytes, of the original APK (8.2KB, 1.4MB, and 45MB, respectively).
Scenario 2 corresponds to the number $n$ of future steps to predict (namely, 1, 12, and 23). 
In scenario 3, it is the size, in bytes, of the original input image (342KB, 1.4MB, and 18MB). 
Finally, in scenario 4 it corresponds to the length of the original input sequence (7, 51 and 178).

We further analyze and discuss these results in the remainder of this section, highlighting the deviation between runs for each payload size considered.

\begin{figure*}[!t]
	\centering
	\includegraphics[scale=0.6]{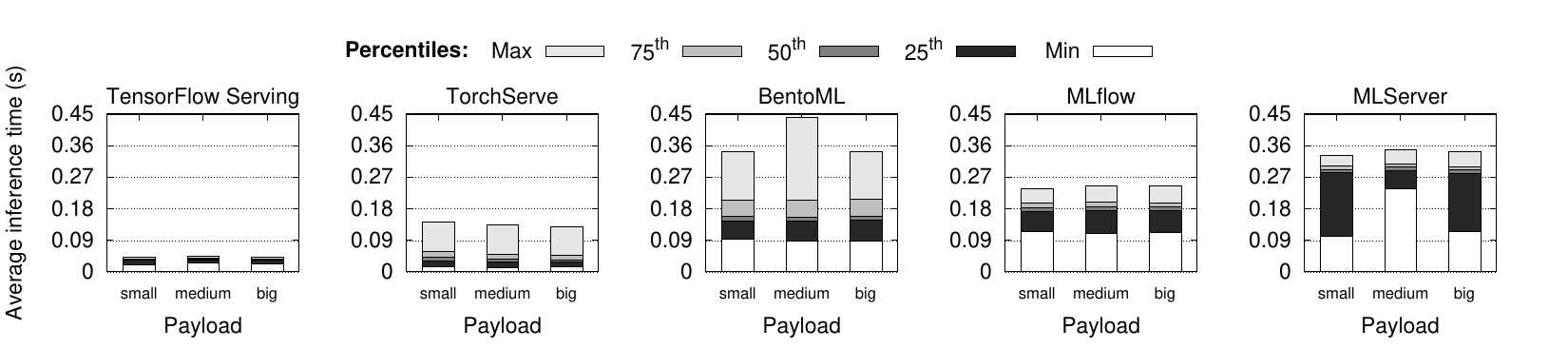}
	\caption{Stacked percentile chart of the average inference time for the serving frameworks in Scenario 1.} 
	\label{fig:scenario1}
\end{figure*}

\begin{figure*}[t]
	\centering
	\includegraphics[scale=0.6]{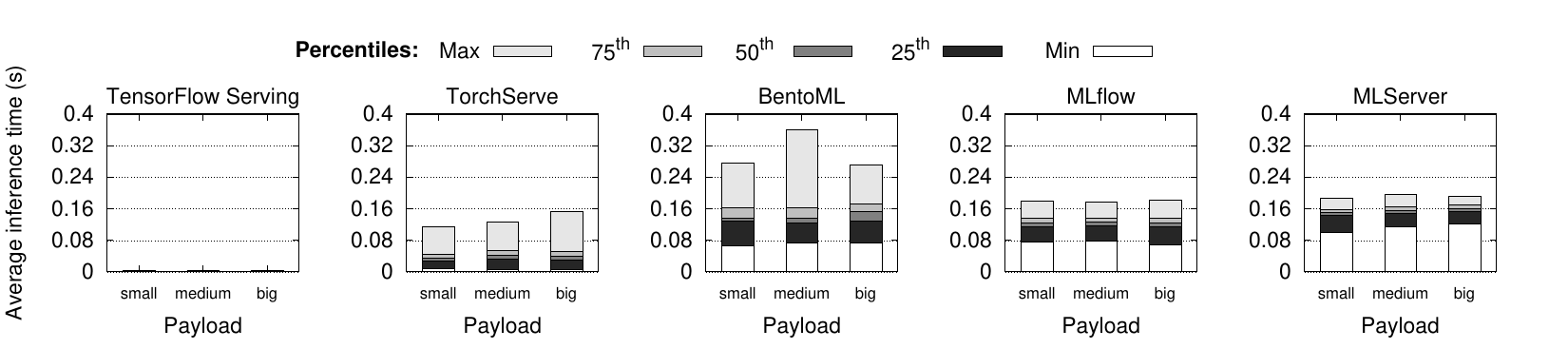}
	\caption{Stacked percentile chart of the average inference time for the serving frameworks in Scenario 2.}
	\label{fig:scenario2}
\end{figure*}

\begin{figure*}[t]
	\centering
	\includegraphics[scale=0.6]{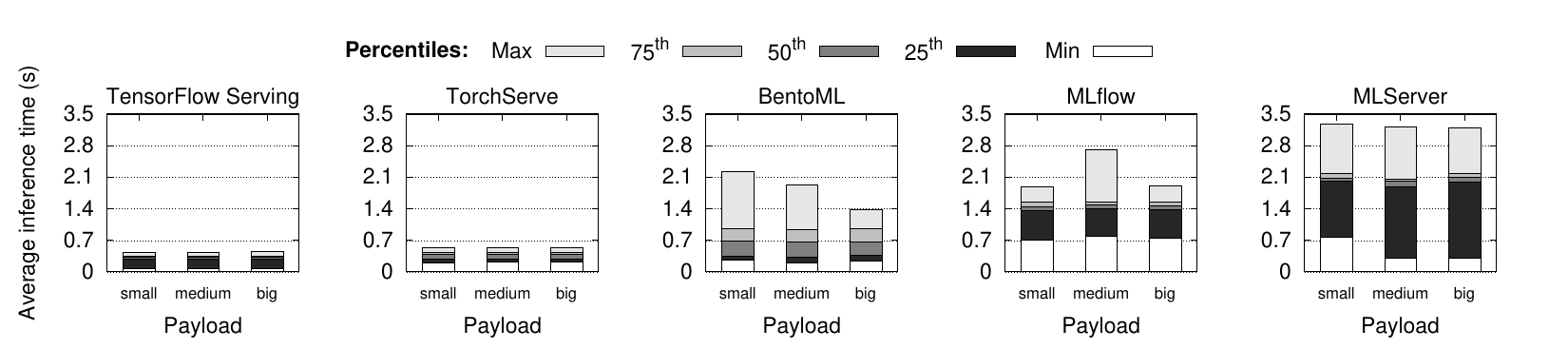}
	\caption{Stacked percentile chart of the average inference time for the serving frameworks in Scenario 3.}
	\label{fig:scenario3}
\end{figure*}

\begin{figure*}[t]
	\centering
	\includegraphics[scale=0.6]{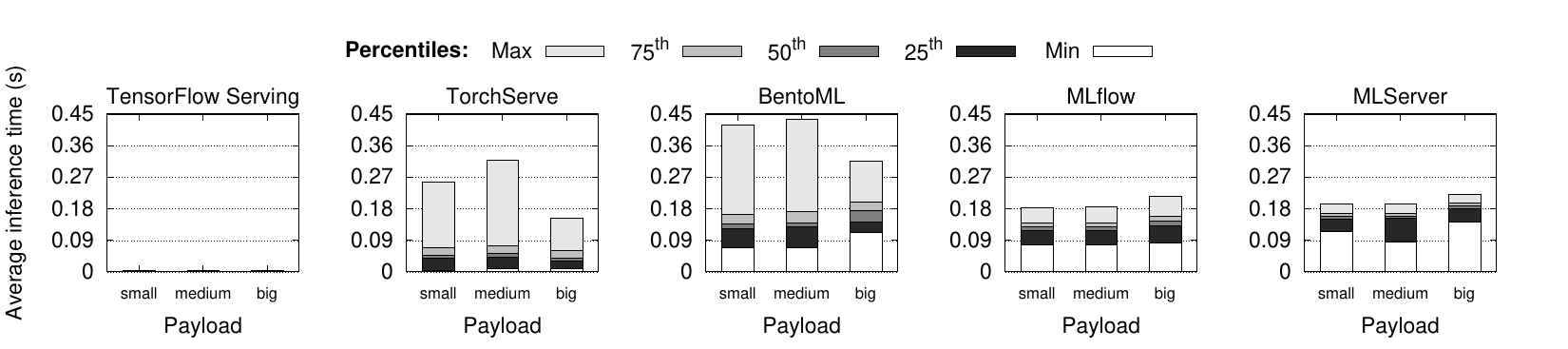}
	\caption{Stacked percentile chart of the average inference time for the serving frameworks in Scenario 4.}
	\label{fig:scenario4}
\end{figure*}

\subsection{Analysis of the results}
\label{sec:discussion}
\textbf{Malware detection}. 
The average number of processed requests in one run (among the two DL-specific frameworks for the three experiments) was 75150, while for the three general-purpose frameworks was 14202.

\begin{tcolorbox}
A1/A4: In this scenario, the two DL-specific scenarios were able to significantly outperform MLflow, MLServer and BentoML. 
\end{tcolorbox}
The minimum average latency (among the two frameworks for the three payloads) was 0.0372s, that is 68.37\% lower than the one observed for the three general-purpose frameworks (0.1176s); on the other hand, the maximum average latency was 0.0895s, 72.04\% lower than the benchmark general-purpose platforms (0.3201s).

TensorFlow Serving showed minimum inference times of, respectively,  0.0215s (small payload), 0.0260s (medium payload) and 0.0242s (high payload), and with maximum latencies (at the 99th percentile) of 0.0431s, 0.0445s, 0.0432s. 

TorchServe showed minimum inference times of, respectively,  0.0136s (small payload), 0.0127s (medium payload) and 0.0135s (high payload), and with maximum latencies of 0.1434s, 0.1346s, 0.1283s.

\begin{tcolorbox}
A3: While TorchServe displayed the lower minimum latencies compared to TensorFlow Serving, the latter proved to be more stable among all the percentiles.
\end{tcolorbox}

\begin{tcolorbox}
A2: No significative differences were detected when changing the payload sizes.
\end{tcolorbox}
	
The only exception was BentoML, that showed an increase in the maximum latency for the medium-sized APK compared to the small and big payloads. 
This shift was only detectable for the 99$^{th}$ percentile, and we believe it was likely due to outliers.\footnote{According to \cite{p9latency}, the variability of response time that leads to high tail latency in a service (expecially at the 99$^{th}$ percentile) can arise for many reasons, including: shared resources, daemons, maintenance activities, queueing, power limits, garbage collection and energy management.}
The results are shown in \autoref{fig:scenario1}.

\textbf{Cryptocoin price forecasting}.
The average number of processed requests in one run (among the two DL-specific frameworks for the three experiments) was 528794, while for the three general-purpose frameworks was 20605.

\begin{tcolorbox}
A1/A4: In this scenario, the two DL-specific scenarios clearly outperform MLflow, MLServer and BentoML. 
\end{tcolorbox}

The minimum average latency (among the two frameworks for the three payloads) was 0.0039s, that is 95.47\% lower than the one observed for the three general-purpose frameworks (0.0860s); on the other hand, the maximum average latency was 0.0676s, 69.95\% lower than the benchmark general-purpose platforms (0.2250s).

TensorFlow Serving showed minimum inference times of, respectively,  0.0008s (small payload), 0.0015s (medium payload) and 0.0016s (high payload), and with maximum latencies (at the 99$^{th}$ percentile) of 0.0035s, 0.0042s, 0.0042s. 

TorchServe showed minimum inference times of, respectively,  0.0093s (small payload), 0.0049s (medium payload) and 0.0054s (high payload), and with maximum latencies of 0.1137s, 0.1262s, 0.1538s.

\begin{tcolorbox}
A3: In this case, TensorFlow Serving showed both lower latencies overall and stability among all the percentiles than TorchServe.
\end{tcolorbox}

\begin{tcolorbox}
A2: No significative differences where detected when changing the payload sizes. 
\end{tcolorbox}

An exception is represented by BentoML and TorchServe, that showed an increase in the maximum latency for the medium-sized input (BentoML) and for the big-sized input (TorchServe) compared to the other payloads, likely due to outliers.\footnote{See note 12.}
The results are shown in \autoref{fig:scenario2}.

\textbf{Image classification}.
The average number of processed requests in one run (among the two DL-specific frameworks for the three experiments) was 8778, while for the three general-purpose frameworks was 2526.
 
\begin{tcolorbox}
A1/A4: The two DL-specific scenarios outperform MLflow, MLServer and BentoML. 
\end{tcolorbox}

The minimum average latency (among the two frameworks for the three payloads) was 0.1427s, that is 70.2\% lower than the one observed for the three general-purpose frameworks (0.4788s); on the other hand, the maximum average latency was 0.4828s, 80.06\% lower than the benchmark general-purpose platforms (2.4213s).

TensorFlow Serving showed minimum inference times of, respectively,  0.0706s (small payload), 0.0756s (medium payload) and 0.0712s (high payload), and with maximum latencies (at the 99$^{th}$ percentile) of 0.4250s, 0.4260s, 0.4444s. 

TorchServe showed minimum inference times of, respectively,  0.2065s (small payload), 0.2188s (medium payload) and 0.2140s (high payload), and with maximum latencies of 0.5357s, 0.5384s, 0.5272s.

\begin{tcolorbox}
A3: In this case, TensorFlow Serving showed lower latencies overall than TorchServe. Both frameworks displayed a general instability among all the percentiles.
\end{tcolorbox}

\begin{tcolorbox}
A2: No significative differences where detected when changing the payload sizes. 
\end{tcolorbox}

An exception is represented by BentoML and MLflow, that showed an increase in the maximum latency for the small-sized input (BentoML) and for the medium-sized input (both BentoML and MLflow) compared to the other payloads. As for the previous scenarios, also in this case the shift is detectable only for the 99$^{th}$ percentile, most likely due to the presence of outliers.\footnote{See note 12.}
The results are shown in \autoref{fig:scenario3}.

\textbf{Sentiment Analysis.}
The average number of processed requests in one run (among the two DL-specific frameworks for the three experiments) was 545141, while for the three general-purpose frameworks was 19162.

\begin{tcolorbox}
A1/A4: As in the other scenarios, the DL-specific platforms outperform MLflow, MLServer and BentoML. 
\end{tcolorbox}

The minimum average latency (among the two frameworks for the three payloads) was 0.0042s, that is 95.44\% lower than the one observed for the three general-purpose frameworks (0.0921s); on the other hand, the maximum average latency was 0.1231s, 53.19\% lower than the benchmark general-purpose platforms (0.2630s).

TensorFlow Serving showed minimum inference times of, respectively,  0.0013s (small payload), 0.0015s (medium payload) and 0.0023s (high payload), and with maximum latencies (at the 99$^{th}$ percentile) of 0.0031s, 0.0038s, 0.0046s. 

TorchServe showed minimum inference times of, respectively,  0.0032s (small payload), 0.0087s (medium payload) and 0.0085s (high payload), and with maximum latencies of 0.2558s, 0.3177s, 0.1541s.

\begin{tcolorbox}
A3: In this case, TensorFlow Serving showed both lower latencies overall and stability among all the percentiles than TorchServe.
\end{tcolorbox}

\begin{tcolorbox}
A2: No significative differences where detected when changing the payload sizes.
\end{tcolorbox}
 
An exception is represented by BentoML and TorchServe, that showed an increase in the maximum latency for the small and medium-sized input compared to the other payloads. 
The results are shown in \autoref{fig:scenario4}.

\subsection{Lessons learned}
\label{sec:lessons}
In all the considered scenarios, TensorFlow Serving results clearly to be the best choice, outperforming every other framework.

A possible reason for this behavior is that the four models have been developed in TensorFlow. In fact, TensorFlow Serving is specifically designed and optimized for serving TensorFlow models, being able to leverage the framework’s full capabilities and providing native support for SavedModels formats. 

The second-best choice results to be TorchServe. Also in this case, it is safe to assume that this is due to the framework being specifically developed to serve PyTorch models. As stated in \autoref{sec:eval}, in fact, in order to evaluate this framework we needed to convert the original TensorFlow model in a PyTorch pretrained format (\texttt{.pth}). This resulting model contains the same weights and parameters of the original one, and therefore provides the same predictions where evaluated.

In the comparison between TensorFlow Serving and TorchServe, we can notice in the second a more unstable behavior, that reflects in outlier high peaks in latency as displayed in \autoref{fig:scenario1}, \autoref{fig:scenario2} and \autoref{fig:scenario4}, despite a general low-latency trend shown in the remaining quartiles.

Overall, the two DL-specific frameworks (TensorFlow Serving and TorchServe) are able to considerably outperform the three general-purpose ML frameworks (BentoML, MLFlow and MLServer) in all the considered scenarios. 

Therefore, we can conclude that when serving DL pretrained models, TensorFlow Serving (for TensorFlow models) and TorchServe (for PyTorch models) are the most efficient solutions available open source to date.

\subsection{Macro-benchmark: turn-around inference time}
In this section, we describe a macro-benchmark analysis conducted among the five serving frameworks in order to describe the turn-around inference time in three selected scenarios. Our aim was to simulate a full DL model serving pipeline, starting from a user request containing the raw, unprocessed input data, to the final postprocessed inference in human-readable format. We leverage Flask \cite{flask}, a Python web framework designed to build lightweight applications, to put in place a local server able to, in order: \emph{(i)} receive a user HTTP request, \emph{(ii)} preprocess the input data, \emph{(iii)} send an inference request to the corresponding ML serving API, and finally \emph{(iv)} send back the resulting, post-processed predictions in human-readable format (\eg labels or classes). 

In the following paragraphs, we briefly describe the main results observed for each scenario.

\begin{figure*}[!t]
	\centering
	\includegraphics[scale=0.6]{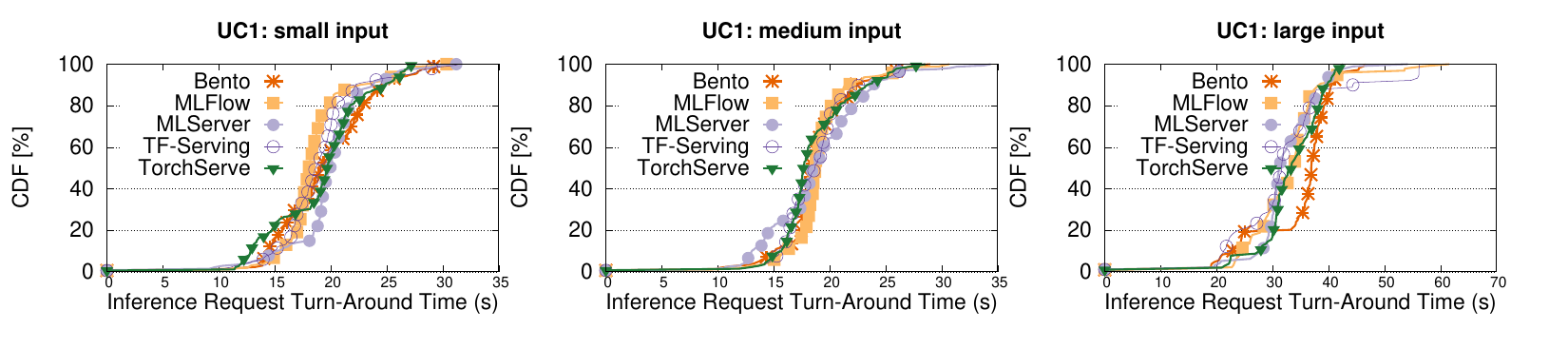}
	\caption{Cumulative distribution function of the request turn-around time for the serving frameworks in Scenario 1.} 
	\label{fig:scenario1:flask}
\end{figure*}

\textbf{Malware detection}. 
As we can observe from \autoref{fig:scenario1:flask}, the average time to process a full request, for each APK size, was significally high. More specifically, each framework suffered from a "cold start", due to an elevated time spent in the features extraction from the input application. No noticeable differences could be observed between the single frameworks latencies. TensorFlow Serving and TorchServe performed slightly better than the other considered frameworks, mostly due to a shorter average inference time (as we discussed in \autoref{sec:eval}). No strong differences in the turn-around time were detected for the small and medium APK, while the larger input presented significally higher latencies due mostly to the massive pre-processing involved.

\begin{figure*}[t]
	\centering
	\includegraphics[scale=0.6]{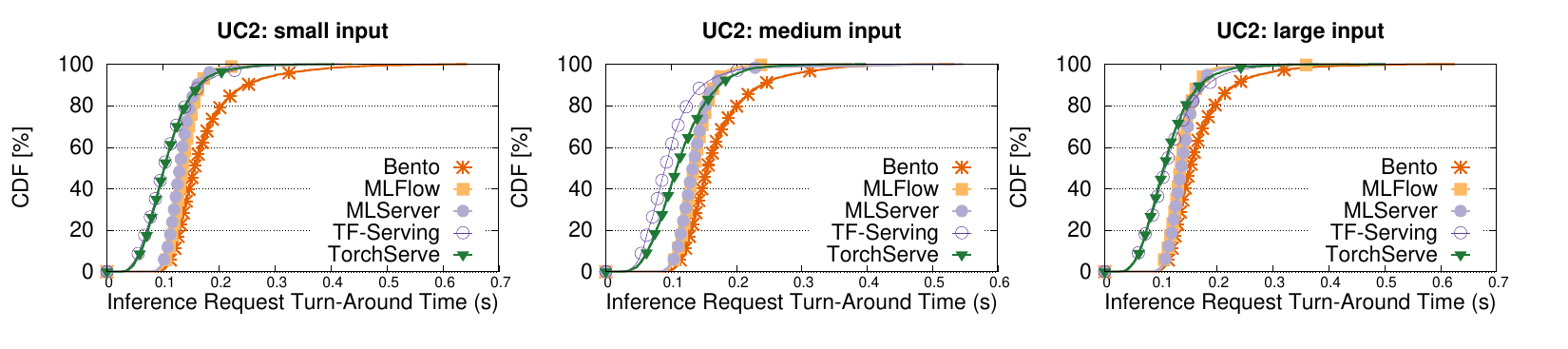}
	\caption{Cumulative distribution function of the request turn-around time for the serving frameworks in Scenario 2.}
	\label{fig:scenario2:flask}
\end{figure*}

\textbf{Cryptocoin price forecasting}. 
From \autoref{fig:scenario2:flask}, we observe that the average time to process a full request, for each number of forecasting steps, was short. No noticeable differences could be observed between the single frameworks latencies. Also here, TensorFlow Serving and TorchServe performed slightly better than the other considered frameworks.
No noticeable differences could be detected for the different input sizes, that did not effect the request turn-around time. This was mostly due to the "lightweight" nature of the input, that did not involve complex operations of pre and post processing.

\begin{figure*}[t]
	\centering
	\includegraphics[scale=0.6]{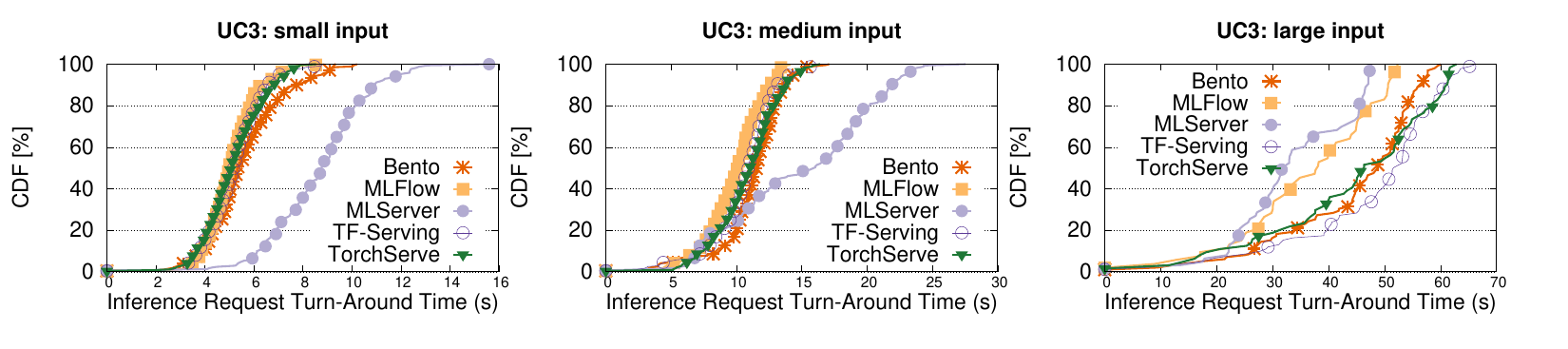}
	\caption{Cumulative distribution function of the request turn-around time for the serving frameworks in Scenario 3.}
	\label{fig:scenario3:flask}
\end{figure*}

\textbf{Image classification}. 
From \autoref{fig:scenario3:flask}, we observe a similar behavior to the first scenario. Also in this case, in fact, the input images in JPEG format could require a significant pre-processing time, that reflected in a "cold start" for all the frameworks. No noticeable differences could be observed between the single frameworks latencies, with the only exception of MLServer. 
In this case, we could detect some differences in the turn-around time with respect to the input sizes, with the larger JPEG image that displayed very high latencies due to the massive pre-processing involved for it.


\section{Related work}
\label{sec:rw}
The study of model serving strategies in ML development and MLOps has recently attracted the interest of academia.

In \cite{relwork1}, the authors present the first performance evaluation study of model serving integration tools in stream processing frameworks. They leveraged Apache Flink as a representative stream processing system and evaluated DL serving pipelines for the image classification task. The study results indicated superior throughput for pipelines that use
embedded libraries to serve pre-trained models. Moreover, in the comparison between TorchServe and TensorFlow
Serving, the latter consistently exhibited better performance, a behavior also confirmed in our paper.

Other recent works mainly focused on efficient and scalable serving techniques for large language models (LLMs). For example, \cite{relwork4} introduced SpecInfer, a system that accelerates generative LLM serving with tree-based speculative inference and verification. It leverages small speculative models to predict the LLM's outputs and organizes the resulting predictions as a token tree, whose nodes each represent a candidate token sequence. It was able to reduce the end-to-end latency while preserving model quality significantly.

In \cite{relwork3}, the authors propose Déjà Vu, a system targeted to reduce Distributed LLM serving inner costs. It implements efficient prompt-token disaggregation to reduce pipeline bubbles, micro batch swapping for efficient GPU memory management, and state replication for fault tolerance.

Finally, \cite{relwork2} introduced the first real-world trace dataset of LLM serving workloads, detailing user, system, and LLM behaviors. Moreover, by analyzing these traces, the authors developed a benchmark suite to enable the performance evaluation of different serving systems.

\section{Conclusion}
\label{sec:conclusion}
Model serving strategies are crucial in deploying ML models. They ensure that the models can be accessed by end-users and utilized efficiently in real-world applications. 
In this paper, we provided a comparison between five state-of-the-art ML model serving frameworks (TensorFlow-Serving, TorchServe, MLServer, MLflow, and BentoML) on four real-world applicative scenarios (malware detection, cryptocoin prices forecasting, image classification, and sentiment analysis). More specifically, we analyzed different strategies of deep learning (DL) model serving under realistic circumstances and assessed which framework is most efficient regarding latencies. 
Our study demonstrated that TensorFlow Serving can outperform every other framework in all the considered scenarios, mostly due to its architecture and structure, which are defined to efficiently serve TensorFlow models. 
Moreover, the two DL-specific frameworks (TensorFlow Serving and TorchServe) achieved significantly lower latencies than the three general-purpose ML frameworks (BentoML, MLFlow, and MLServer), proving to be the two most efficient open-source solutions for serving DL models.

Finally, we aim to extend our work to real-world applications where other cost measures, different from latency, might be impactful. One is surely confidential ML, where models have to be served and deployed in a manner that ensures data privacy and model security. This is especially crucial in domains like healthcare, finance, and defense, where the confidentiality of data and models is paramount. Moreover, we plan to further enrich our analysis by considering also models trained using frameworks other than TensorFlow, like PyTorch, JAX \cite{jax2018github} and Caffe \cite{jia2014caffe}.



\bibliographystyle{plain}
\bibliography{biblio}

\begin{thebibliography}{10}

\bibitem{aws}
Amazon sagemaker website.
\newblock \url{https://aws.amazon.com/sagemaker/pricing/} (accessed July 11,
  2024).

\bibitem{realworld3}
Asticaai website.
\newblock \url{https://astica.ai/vision/} (accessed July 11, 2024).

\bibitem{azure}
Azure machine learning website.
\newblock
  \url{https://azure.microsoft.com/en-us/pricing/details/machine-learning/}
  (accessed July 11, 2024).

\bibitem{bentoml}
Bentoml documentation.
\newblock \url{https://docs.bentoml.com/en/latest/} (accessed July 11, 2024).

\bibitem{binance}
Binance documentation.
\newblock \url{https://www.binance.com/} (accessed July 11, 2024).

\bibitem{catboost}
Catboost documentation.
\newblock \url{https://catboost.ai/} (accessed July 11, 2024).

\bibitem{contagio}
Contagio security blog.
\newblock \url{https://contagiodump.blogspot.com/} (accessed July 11, 2024).

\bibitem{flask}
Flask documentation.
\newblock \url{https://flask.palletsprojects.com/en/3.0.x/} (accessed July 11,
  2024).

\bibitem{googleai}
Google cloud ai website.
\newblock \url{https://cloud.google.com/ai-platform/pricing} (accessed July 11,
  2024).

\bibitem{hfts}
Huggingface and torchserve integration.
\newblock
  \url{https://github.com/pytorch/serve/blob/master/examples/Huggingface_Transformers/Transformer_handler_generalized.py}
  (accessed July 11, 2024).

\bibitem{ilsvrc12}
Ilsvrc 2012 website.
\newblock \url{https://www.image-net.org/challenges/LSVRC/2012/} (accessed July
  11, 2024).

\bibitem{realworld1}
Kaspersky cloud ml for android.
\newblock
  \url{https://usa.kaspersky.com/about/press-releases/2019_cloud-ml-for-android}
  (accessed July 11, 2024).

\bibitem{kerasa}
Keras applications documentation.
\newblock \url{https://keras.io/api/applications/} (accessed July 11, 2024).

\bibitem{kserve}
Kserve documentation.
\newblock \url{https://kserve.github.io/website/latest/} (accessed July 11,
  2024).

\bibitem{kubeflow}
Kubeflow documentation.
\newblock \url{https://www.kubeflow.org/} (accessed July 11, 2024).

\bibitem{kubernetes}
Kubernetes documentation.
\newblock \url{https://kubernetes.io/docs/home/} (accessed July 11, 2024).

\bibitem{lightgbm}
Lightgbm documentation.
\newblock \url{https://lightgbm.readthedocs.io/en/stable/} (accessed July 11,
  2024).

\bibitem{mar}
Mar file specifics.
\newblock \url{https://pytorch.org/serve/FAQs.html#what-is-a-mar-file}
  (accessed July 11, 2024).

\bibitem{mlflow}
Mlflow documentation.
\newblock \url{https://mlflow.org/docs/latest/index.html} (accessed July 11,
  2024).

\bibitem{mlserver}
Mlserver documentation.
\newblock \url{https://mlserver.readthedocs.io/en/stable/} (accessed July 11,
  2024).

\bibitem{oha}
Oha documentation.
\newblock \url{https://github.com/hatoo/oha} (accessed July 11, 2024).

\bibitem{onnx}
Onnx documentation.
\newblock \url{https://onnx.ai/onnx/index.html} (accessed July 11, 2024).

\bibitem{onnx2torch}
Onnx2torch documentation.
\newblock \url{https://github.com/ENOT-AutoDL/onnx2torch} (accessed July 11,
  2024).

\bibitem{sklearn}
Scikit-learn documentation.
\newblock \url{https://scikit-learn.org/stable/} (accessed July 11, 2024).

\bibitem{tflite}
Tensorflow lite documentation.
\newblock \url{https://www.tensorflow.org/lite} (accessed July 11, 2024).

\bibitem{tfs}
Tensorflow serving documentation.
\newblock \url{https://www.tensorflow.org/tfx/guide/serving} (accessed July 11,
  2024).

\bibitem{tfjs}
Tensorflow.js documentation.
\newblock \url{https://www.tensorflow.org/js} (accessed July 11, 2024).

\bibitem{realworld4}
Text2data website.
\newblock \url{https://text2data.com/Demo} (accessed July 11, 2024).

\bibitem{tf2onnx}
Tf2onnx documentation.
\newblock \url{https://github.com/onnx/tensorflow-onnx} (accessed July 11,
  2024).

\bibitem{torchscript}
Torchscript documentation.
\newblock \url{https://pytorch.org/docs/stable/jit.html} (accessed July 11,
  2024).

\bibitem{ts}
Torchserve documentation.
\newblock \url{https://pytorch.org/serve/} (accessed July 11, 2024).

\bibitem{virustotal}
Virustotal website.
\newblock \url{https://www.virustotal.com/} (accessed July 11, 2024).

\bibitem{realworld2}
Walletinvestor website.
\newblock \url{https://walletinvestor.com/forecast} (accessed July 11, 2024).

\bibitem{xgboost}
Xgboost documentation.
\newblock \url{https://xgboost.readthedocs.io/en/stable/} (accessed July 11,
  2024).

\bibitem{tensorflow2015-whitepaper}
Mart\'{i}n Abadi, Ashish Agarwal, Paul Barham, Eugene Brevdo, Zhifeng Chen,
  Craig Citro, Greg~S. Corrado, Andy Davis, Jeffrey Dean, Matthieu Devin,
  Sanjay Ghemawat, Ian Goodfellow, Andrew Harp, Geoffrey Irving, Michael Isard,
  Yangqing Jia, Rafal Jozefowicz, Lukasz Kaiser, Manjunath Kudlur, Josh
  Levenberg, Dandelion Man\'{e}, Rajat Monga, Sherry Moore, Derek Murray, Chris
  Olah, Mike Schuster, Jonathon Shlens, Benoit Steiner, Ilya Sutskever, Kunal
  Talwar, Paul Tucker, Vincent Vanhoucke, Vijay Vasudevan, Fernanda Vi\'{e}gas,
  Oriol Vinyals, Pete Warden, Martin Wattenberg, Martin Wicke, Yuan Yu, and
  Xiaoqiang Zheng.
\newblock {TensorFlow}: Large-scale machine learning on heterogeneous systems,
  2015.
\newblock Software available from tensorflow.org.

\bibitem{appl1}
Ryo Akita, Akira Yoshihara, Takashi Matsubara, and Kuniaki Uehara.
\newblock Deep learning for stock prediction using numerical and textual
  information.
\newblock In {\em 2016 IEEE/ACIS 15th International Conference on Computer and
  Information Science (ICIS)}, pages 1--6, 2016.

\bibitem{drebin}
Dan Arp, Michael Spreitzenbarth, Malte Hubner, Hugo Gascon, and Konrad Rieck.
\newblock Drebin: Effective and explainable detection of android malware in
  your pocket.
\newblock In {\em Network and Distributed System Security Symposium}, 2014.

\bibitem{appl7}
Alexei Baevski, Henry Zhou, Abdelrahman Mohamed, and Michael Auli.
\newblock wav2vec 2.0: a framework for self-supervised learning of speech
  representations.
\newblock In {\em Proceedings of the 34th International Conference on Neural
  Information Processing Systems}, NIPS '20, Red Hook, NY, USA, 2020. Curran
  Associates Inc.

\bibitem{appl6}
Alexey Bochkovskiy, Chien{-}Yao Wang, and Hong{-}Yuan~Mark Liao.
\newblock Yolov4: Optimal speed and accuracy of object detection.
\newblock {\em CoRR}, abs/2004.10934, 2020.

\bibitem{crypto1}
Ahmed Bouteska, Mohammad~Zoynul Abedin, Petr Hajek, and Kunpeng Yuan.
\newblock Cryptocurrency price forecasting – a comparative analysis of
  ensemble learning and deep learning methods.
\newblock {\em International Review of Financial Analysis}, 92:103055, 2024.

\bibitem{jax2018github}
James Bradbury, Roy Frostig, Peter Hawkins, Matthew~James Johnson, Chris Leary,
  Dougal Maclaurin, George Necula, Adam Paszke, Jake Vander{P}las, Skye
  Wanderman-{M}ilne, and Qiao Zhang.
\newblock {JAX}: composable transformations of {P}ython+{N}um{P}y programs,
  2018.

\bibitem{appl5}
Nicolas Carion, Francisco Massa, Gabriel Synnaeve, Nicolas Usunier, Alexander
  Kirillov, and Sergey Zagoruyko.
\newblock End-to-end object detection with transformers.
\newblock In Andrea Vedaldi, Horst Bischof, Thomas Brox, and Jan-Michael Frahm,
  editors, {\em Computer Vision -- ECCV 2020}, pages 213--229, Cham, 2020.
  Springer International Publishing.

\bibitem{appl8}
William Chan, Navdeep Jaitly, Quoc Le, and Oriol Vinyals.
\newblock Listen, attend and spell: A neural network for large vocabulary
  conversational speech recognition.
\newblock In {\em 2016 IEEE International Conference on Acoustics, Speech and
  Signal Processing (ICASSP)}, pages 4960--4964, 2016.

\bibitem{gru}
Kyunghyun Cho, Bart van Merri{\"e}nboer, Dzmitry Bahdanau, and Yoshua Bengio.
\newblock On the properties of neural machine translation: Encoder{--}decoder
  approaches.
\newblock In {\em Proceedings of {SSST}-8, Eighth Workshop on Syntax, Semantics
  and Structure in Statistical Translation}, pages 103--111, Doha, Qatar, oct
  2014. Association for Computational Linguistics.

\bibitem{clipper}
Daniel Crankshaw, Xin Wang, Giulio Zhou, Michael~J. Franklin, Joseph~E.
  Gonzalez, and Ion Stoica.
\newblock Clipper: a low-latency online prediction serving system.
\newblock In {\em Proceedings of the 14th USENIX Conference on Networked
  Systems Design and Implementation}, NSDI'17, page 613–627, USA, 2017.
  USENIX Association.

\bibitem{crypto3}
Pasquale De~Rosa, Pascal Felber, and Valerio Schiavoni.
\newblock Practical forecasting of cryptocoins timeseries using correlation
  patterns.
\newblock In {\em Proceedings of the 17th ACM International Conference on
  Distributed and Event-Based Systems}, DEBS '23, page 80–90, New York, NY,
  USA, 2023. Association for Computing Machinery.

\bibitem{p9latency}
Jeffrey Dean and Luiz~Andr\'{e} Barroso.
\newblock The tail at scale.
\newblock {\em Commun. ACM}, 56(2):74–80, feb 2013.

\bibitem{sentiment1}
Zulfadzli Drus and Haliyana Khalid.
\newblock Sentiment analysis in social media and its application: Systematic
  literature review.
\newblock {\em Procedia Computer Science}, 161:707--714, 2019.
\newblock The Fifth Information Systems International Conference, 23-24 July
  2019, Surabaya, Indonesia.

\bibitem{md1}
Omar~N. Elayan and Ahmad~M. Mustafa.
\newblock Android malware detection using deep learning.
\newblock {\em Procedia Computer Science}, 184:847--852, 2021.
\newblock The 12th International Conference on Ambient Systems, Networks and
  Technologies (ANT) / The 4th International Conference on Emerging Data and
  Industry 4.0 (EDI40) / Affiliated Workshops.

\bibitem{lstm}
Sepp Hochreiter and J\"{u}rgen Schmidhuber.
\newblock Long short-term memory.
\newblock {\em Neural Comput.}, 9(8):1735–1780, nov 1997.

\bibitem{relwork1}
Sonia Horchidan, Emmanouil Kritharakis, Vasiliki Kalavri, and Paris Carbone.
\newblock Evaluating model serving strategies over streaming data.
\newblock In {\em Proceedings of the Sixth Workshop on Data Management for
  End-To-End Machine Learning}, DEEM '22, New York, NY, USA, 2022. Association
  for Computing Machinery.

\bibitem{staticvsdynamic}
Muhammad Ijaz, Muhammad~Hanif Durad, and Maliha Ismail.
\newblock Static and dynamic malware analysis using machine learning.
\newblock In {\em 2019 16th International Bhurban Conference on Applied
  Sciences and Technology (IBCAST)}, pages 687--691, 2019.

\bibitem{crypto2}
Patrick Jaquart, Sven Köpke, and Christof Weinhardt.
\newblock Machine learning for cryptocurrency market prediction and trading.
\newblock {\em The Journal of Finance and Data Science}, 8:331--352, 2022.

\bibitem{jia2014caffe}
Yangqing Jia, Evan Shelhamer, Jeff Donahue, Sergey Karayev, Jonathan Long, Ross
  Girshick, Sergio Guadarrama, and Trevor Darrell.
\newblock Caffe: Convolutional architecture for fast feature embedding.
\newblock {\em arXiv preprint arXiv:1408.5093}, 2014.

\bibitem{joseph}
V.~Roshan Joseph.
\newblock Optimal ratio for data splitting.
\newblock {\em Statistical Analysis and Data Mining: The ASA Data Science
  Journal}, 15(4):531–538, April 2022.

\bibitem{convnetcost}
Alex Krizhevsky, Ilya Sutskever, and Geoffrey~E. Hinton.
\newblock Imagenet classification with deep convolutional neural networks.
\newblock {\em Commun. ACM}, 60(6):84–90, may 2017.

\bibitem{cnn}
Yann LeCun and Yoshua Bengio.
\newblock Convolutional networks for images, speech, and time series.
\newblock In Michael~A. Arbib, editor, {\em Handbook of Brain Theory and Neural
  Networks}, page 3361. MIT Press, 1995.

\bibitem{md2}
Kaijun Liu, Shengwei Xu, Guoai Xu, Miao Zhang, Dawei Sun, and Haifeng Liu.
\newblock A review of android malware detection approaches based on machine
  learning.
\newblock {\em IEEE Access}, 8:124579--124607, 2020.

\bibitem{imdb}
Andrew~L. Maas, Raymond~E. Daly, Peter~T. Pham, Dan Huang, Andrew~Y. Ng, and
  Christopher Potts.
\newblock Learning word vectors for sentiment analysis.
\newblock In {\em Proceedings of the 49th Annual Meeting of the Association for
  Computational Linguistics: Human Language Technologies}, pages 142--150,
  Portland, Oregon, USA, June 2011. Association for Computational Linguistics.

\bibitem{mald20}
Samaneh Mahdavifar, Andi~Fitriah Abdul~Kadir, Rasool Fatemi, Dima Alhadidi, and
  Ali~A. Ghorbani.
\newblock Dynamic android malware category classification using semi-supervised
  deep learning.
\newblock In {\em 2020 IEEE Intl Conf on Dependable, Autonomic and Secure
  Computing, Intl Conf on Pervasive Intelligence and Computing, Intl Conf on
  Cloud and Big Data Computing, Intl Conf on Cyber Science and Technology
  Congress (DASC/PiCom/CBDCom/CyberSciTech)}, pages 515--522, 2020.

\bibitem{image1}
Pawan~Kumar Mall, Pradeep~Kumar Singh, Swapnita Srivastav, Vipul Narayan,
  Marcin Paprzycki, Tatiana Jaworska, and Maria Ganzha.
\newblock A comprehensive review of deep neural networks for medical image
  processing: Recent developments and future opportunities.
\newblock {\em Healthcare Analytics}, 4:100216, 2023.

\bibitem{relwork4}
Xupeng Miao, Gabriele Oliaro, Zhihao Zhang, Xinhao Cheng, Zeyu Wang, Zhengxin
  Zhang, Rae Ying~Yee Wong, Alan Zhu, Lijie Yang, Xiaoxiang Shi, Chunan Shi,
  Zhuoming Chen, Daiyaan Arfeen, Reyna Abhyankar, and Zhihao Jia.
\newblock Specinfer: Accelerating large language model serving with tree-based
  speculative inference and verification.
\newblock In {\em Proceedings of the 29th ACM International Conference on
  Architectural Support for Programming Languages and Operating Systems, Volume
  3}, ASPLOS '24, page 932–949, New York, NY, USA, 2024. Association for
  Computing Machinery.

\bibitem{wordnet}
George~A. Miller.
\newblock {W}ord{N}et: A lexical database for {E}nglish.
\newblock In {\em {H}uman {L}anguage {T}echnology: Proceedings of a Workshop
  held at {P}lainsboro, {N}ew {J}ersey, {M}arch 8-11, 1994}, 1994.

\bibitem{appl2}
Latrisha~N. Mintarya, Jeta~N.M. Halim, Callista Angie, Said Achmad, and Aditya
  Kurniawan.
\newblock Machine learning approaches in stock market prediction: A systematic
  literature review.
\newblock {\em Procedia Computer Science}, 216:96--102, 2023.
\newblock 7th International Conference on Computer Science and Computational
  Intelligence 2022.

\bibitem{rnnvisual}
Volodymyr Mnih, Nicolas Heess, Alex Graves, and Koray Kavukcuoglu.
\newblock Recurrent models of visual attention.
\newblock In {\em Proceedings of the 27th International Conference on Neural
  Information Processing Systems - Volume 2}, NIPS'14, page 2204–2212,
  Cambridge, MA, USA, 2014. MIT Press.

\bibitem{olston2017tensorflow}
Christopher Olston, Noah Fiedel, Kiril Gorovoy, Jeremiah Harmsen, Li~Lao,
  Fangwei Li, Vinu Rajashekhar, Sukriti Ramesh, and Jordan Soyke.
\newblock {Tensorflow-serving: Flexible, high-performance ml serving}.
\newblock {\em arXiv preprint arXiv:1712.06139}, 2017.

\bibitem{conf1}
Nicolas Papernot, Patrick McDaniel, Arunesh Sinha, and Michael~P. Wellman.
\newblock Sok: Security and privacy in machine learning.
\newblock In {\em 2018 IEEE European Symposium on Security and Privacy}, pages
  399--414, 2018.

\bibitem{torch}
Adam Paszke, Sam Gross, Francisco Massa, Adam Lerer, James Bradbury, Gregory
  Chanan, Trevor Killeen, Zeming Lin, Natalia Gimelshein, Luca Antiga, Alban
  Desmaison, Andreas K\"{o}pf, Edward Yang, Zach DeVito, Martin Raison, Alykhan
  Tejani, Sasank Chilamkurthy, Benoit Steiner, Lu~Fang, Junjie Bai, and Soumith
  Chintala.
\newblock {\em PyTorch: an imperative style, high-performance deep learning
  library}.
\newblock Curran Associates Inc., Red Hook, NY, USA, 2019.

\bibitem{mlaas}
Robert Philipp, Andreas Mladenow, Christine Strauss, and Alexander V\"{o}lz.
\newblock Machine learning as a service: Challenges in research and
  applications.
\newblock In {\em Proceedings of the 22nd International Conference on
  Information Integration and Web-Based Applications \& Services}, iiWAS '20,
  page 396–406, New York, NY, USA, 2021. Association for Computing Machinery.

\bibitem{image2}
Joseph Redmon, Santosh Divvala, Ross Girshick, and Ali Farhadi.
\newblock You only look once: Unified, real-time object detection.
\newblock In {\em 2016 IEEE Conference on Computer Vision and Pattern
  Recognition (CVPR)}, pages 779--788, 2016.

\bibitem{ribeiro2015mlaas}
Mauro Ribeiro, Katarina Grolinger, and Miriam~AM Capretz.
\newblock {Mlaas: Machine learning as a service}.
\newblock In {\em 2015 IEEE 14th international conference on machine learning
  and applications (ICMLA)}, pages 896--902. IEEE, 2015.

\bibitem{imagenet}
Olga Russakovsky, Jia Deng, Hao Su, Jonathan Krause, Sanjeev Satheesh, Sean Ma,
  Zhiheng Huang, Andrej Karpathy, Aditya Khosla, Michael Bernstein,
  Alexander~C. Berg, and Li~Fei-Fei.
\newblock {ImageNet Large Scale Visual Recognition Challenge}.
\newblock {\em International Journal of Computer Vision (IJCV)},
  115(3):211--252, 2015.

\bibitem{md3}
Justin Sahs and Latifur Khan.
\newblock A machine learning approach to android malware detection.
\newblock In {\em 2012 European Intelligence and Security Informatics
  Conference}, pages 141--147, 2012.

\bibitem{mobilenetv2}
Mark Sandler, Andrew Howard, Menglong Zhu, Andrey Zhmoginov, and Liang-Chieh
  Chen.
\newblock Mobilenetv2: Inverted residuals and linear bottlenecks.
\newblock In {\em 2018 IEEE/CVF Conference on Computer Vision and Pattern
  Recognition}, pages 4510--4520, 2018.

\bibitem{cnn2}
Neha Sharma, Vibhor Jain, and Anju Mishra.
\newblock An analysis of convolutional neural networks for image
  classification.
\newblock {\em Procedia Computer Science}, 132:377--384, 2018.
\newblock International Conference on Computational Intelligence and Data
  Science.

\bibitem{conf2}
Sagar Sharma and Keke Chen.
\newblock Confidential machine learning on untrusted platforms: a survey.
\newblock {\em Cybersecurity}, 4, 2020.

\bibitem{sentiment2}
T.~K. Shivaprasad and Jyothi Shetty.
\newblock Sentiment analysis of product reviews: A review.
\newblock In {\em 2017 International Conference on Inventive Communication and
  Computational Technologies (ICICCT)}, pages 298--301, 2017.

\bibitem{relwork3}
Foteini Strati, Sara Mcallister, Amar Phanishayee, Jakub Tarnawski, and Ana
  Klimovic.
\newblock D\'ej\`avu: Kv-cache streaming for fast, fault-tolerant generative
  llm serving, 2024.

\bibitem{appl3}
Fei Sun, Jun Liu, Jian Wu, Changhua Pei, Xiao Lin, Wenwu Ou, and Peng Jiang.
\newblock Bert4rec: Sequential recommendation with bidirectional encoder
  representations from transformer.
\newblock In {\em Proceedings of the 28th ACM International Conference on
  Information and Knowledge Management}, CIKM '19, page 1441–1450, New York,
  NY, USA, 2019. Association for Computing Machinery.

\bibitem{relwork2}
Yuxin Wang, Yuhan Chen, Zeyu Li, Zhenheng Tang, Rui Guo, Xin Wang, Qiang Wang,
  Amelie~Chi Zhou, and Xiaowen Chu.
\newblock Towards efficient and reliable llm serving: A real-world workload
  study, 2024.

\bibitem{cloud1}
Mazin Yousif.
\newblock Intelligence in the cloud – we need a lot of it.
\newblock {\em IEEE Cloud Computing}, 4(6):4--6, 2017.

\bibitem{zaharia2018accelerating}
Matei Zaharia, Andrew Chen, Aaron Davidson, Ali Ghodsi, Sue~Ann Hong, Andy
  Konwinski, Siddharth Murching, Tomas Nykodym, Paul Ogilvie, Mani Parkhe,
  et~al.
\newblock {Accelerating the machine learning lifecycle with MLflow}.
\newblock {\em IEEE Data Eng. Bull.}, 41(4):39--45, 2018.

\bibitem{appl4}
Shuai Zhang, Lina Yao, Aixin Sun, and Yi~Tay.
\newblock Deep learning based recommender system: A survey and new
  perspectives.
\newblock {\em ACM Comput. Surv.}, 52(1), feb 2019.

\end{thebibliography}

\end{document}